\documentclass[11pt]{article}

\usepackage[margin=1in]{geometry}
\usepackage[numbers,compress]{natbib}

\usepackage{microtype}
\usepackage{graphicx}
\usepackage{amsmath,amssymb}

\usepackage{latexsym}

\usepackage[T1]{fontenc}

\usepackage[utf8]{inputenc}

\usepackage{microtype}
\usepackage{array}

\usepackage{inconsolata}

\usepackage{graphicx}

\usepackage{hyperref}
\usepackage{url}

\usepackage{algorithm}
\usepackage{algorithmic}

\usepackage{newfloat}
\usepackage{listings}
\usepackage{graphicx}

\usepackage{url}
\usepackage{xspace}
\usepackage{booktabs}
\usepackage{amsfonts}       %
\usepackage{amsthm,amsmath}
\usepackage{natbib}
\usepackage{enumitem}
\usepackage{multirow}
\usepackage{makecell}
\usepackage[table]{xcolor}
\usepackage{graphicx}
\usepackage{lipsum}
\usepackage{algorithm}
\usepackage{algorithmic}
\usepackage{subfig}
\usepackage{wrapfig}
\usepackage[capitalize]{cleveref}

\usepackage{xspace}
\newcommand{\ourmethod}{{\fontfamily{lmtt}\selectfont \textbf{R2F}}\xspace}
\definecolor{tablehead}{RGB}{121,80,242}
\definecolor{mydarkblue}{rgb}{0,0.08,0.45}
\definecolor{mycite}{cmyk}{0.55,1,0,0.15}
\usepackage[loose]{minitoc}

\doparttoc %
\faketableofcontents %

\newcommand{{\full}}{{{\textsf{\small{FullGrad}}}\xspace}}
\newcommand{{\loras}}{{{\textsf{\small{LoRA\_{sin}}}}\xspace}}
\newcommand{{\loram}}{{{\textsf{\small{LoRA\_{mul}}}}\xspace}}
\newcommand{{\scrub}}{{{\textsf{\small{SCRUB}}}\xspace}}
\newcommand{{\eco}}{{{\textsf{\small{ECO}}}\xspace}}
\newcommand{{\sku}}{{{\textsf{\small{SKU}}}\xspace}}
\newcommand{{\cut}}{{{\textsf{\small{DeepCUT}}}\xspace}}

\DeclareCaptionStyle{ruled}{labelfont=normalfont,labelsep=colon,strut=off} %
\floatstyle{ruled}
\newfloat{listing}{tb}{lst}{}
\floatname{listing}{Listing}
\makeatletter
\newenvironment{fullitemize}
{
\vspace{-1pt}
\begin{itemize}[leftmargin=*]
\setlength{\itemsep}{5pt}
\setlength{\parsep}{-5pt}
\setlength{\parskip}{-3pt}
\setlength{\leftmargin}{-10pt}
}
{
\end{itemize}
\vspace{-1pt}
}
\makeatother

\newcolumntype{I}{!{\vrule width 1pt}}

\title{Recover-to-Forget: Gradient Reconstruction from LoRA for Efficient LLM Unlearning}

\author{
  \textbf{Yezi Liu}\textsuperscript{1} \quad
  \textbf{Hanning Chen}\textsuperscript{1} \quad
  \textbf{Wenjun Huang}\textsuperscript{1} \\
\textbf{Yang Ni}\textsuperscript{2} \quad
  \textbf{Mohsen Imani}\textsuperscript{1} \\
  \textsuperscript{1}University of California, Irvine \quad 
  \textsuperscript{2}Purdue University Northwest \quad \\
\texttt{\{yezil3,hanningc,wenjunh3,m.imani\}@uci.edu} \\
\texttt{yangni@purdue.edu}
}
\date{}

\begin{document}

\maketitle

\begin{abstract}
Unlearning in large foundation models (e.g., LLMs) is essential for enabling dynamic knowledge updates, enforcing data deletion rights, and correcting model behavior. However, existing unlearning methods often require full-model fine-tuning or access to the original training data, which limits their scalability and practicality. In this work, we introduce \textbf{R}ecover-\textbf{to}-\textbf{F}orget ({\ourmethod}), a novel framework for efficient unlearning in LLMs based on reconstructing full-model gradient directions from low-rank LoRA adapter updates. Rather than performing backpropagation through the full model, we compute gradients with respect to LoRA parameters using multiple paraphrased prompts and train a gradient decoder to approximate the corresponding full-model gradients. To ensure applicability to larger or black-box models, the decoder is trained on a proxy model and transferred to target models. We provide a theoretical analysis of cross-model generalization and demonstrate that our method achieves effective unlearning while preserving general model performance. Experimental results demonstrate that {\ourmethod} offers a scalable and lightweight alternative for unlearning in pretrained LLMs without requiring full retraining or access to internal parameters.
\end{abstract}

\section{Introduction}
\begin{wrapfigure}[15]{r}{0.49\textwidth}
    \vspace{-3em}
   {\includegraphics[width=1.0\linewidth]{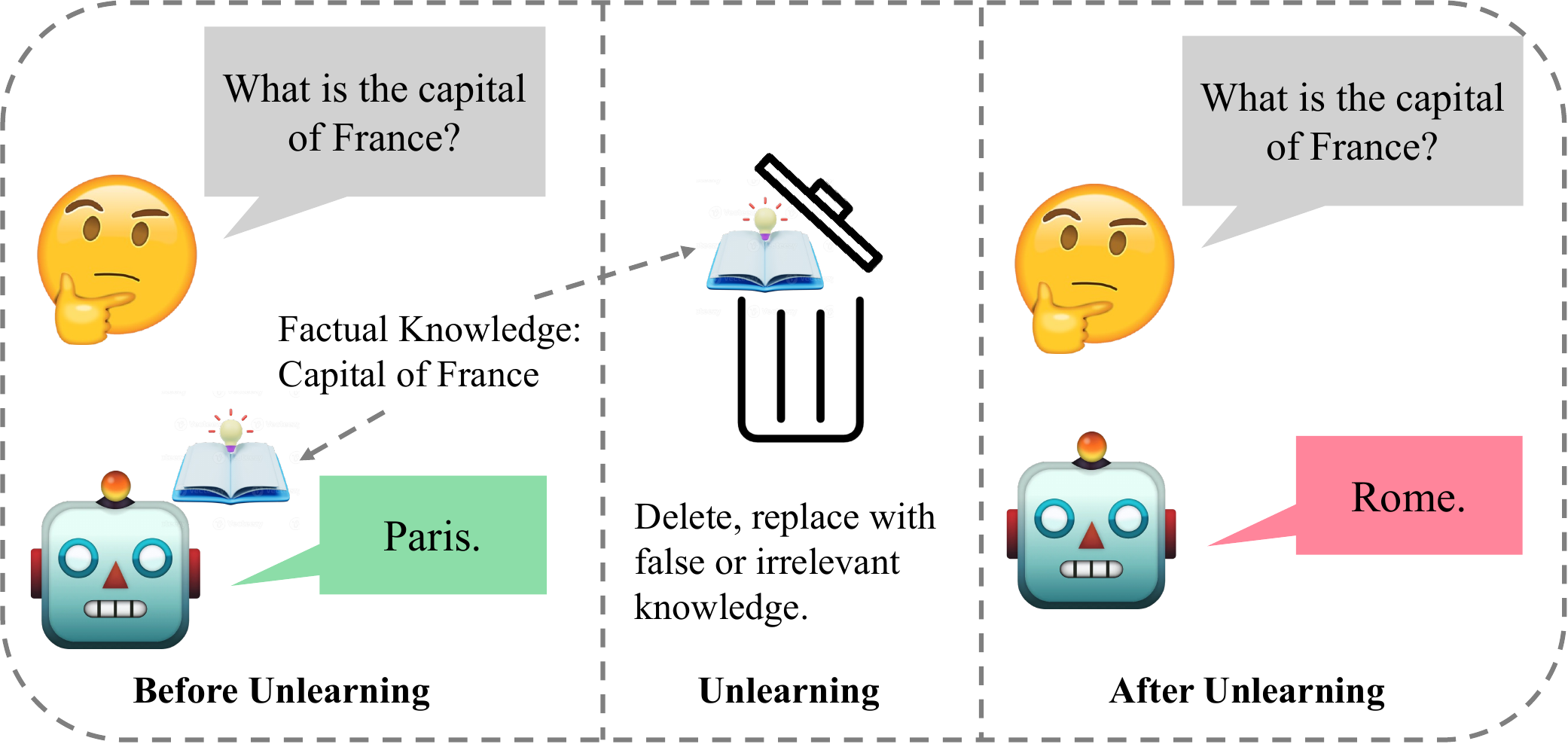}}\label{intro:dp}
    \vspace{-1.5em}
\caption{Illustration of the LLM unlearning task. The model initially answers ``What is the capital of France?'' with ``Paris''. During unlearning, the target fact (``Capital of France $\rightarrow$ Paris'') is removed or corrupted. After unlearning, the model forgets the original answer, responding with incorrect or irrelevant outputs (e.g., ``Rome'') while preserving unrelated knowledge.}
    \vspace{-1em}
    \label{fig:intro}
\end{wrapfigure}    
The widespread deployment of large language models (LLMs), such as GPT-5.1 Thinking~\citep{openai2025gpt51thinking}, Gemini 3.0 Pro~\citep{gemini3pro2025}, and LLaMA 4~\citep{meta2025llama4}, has significantly advanced various natural language processing tasks. Recently, the trustworthiness of large language models (LLMs), including issues such as fairness~\citep{liu2025enabling,liu2025fgu,liu2024promoting,liu2023fairgraph}, safety~\citep{liu2025lune}, and robustness~\citep{liucauchy}, has attracted increasing attention~\citep{liu2025white}. At the same time, their rapid adoption has amplified concerns around data privacy, particularly regarding the inadvertent memorization of sensitive or proprietary information~\citep{carlini2021extracting, jagielski2022measuring}. Such memorization poses risks of data leakage, violating privacy regulations like GDPR, which explicitly advocates for the \textit{right to be forgotten}~\citep{ginart2019making}. Consequently, there is an increasing demand for efficient machine unlearning (MU) techniques capable of selectively removing specific information from LLMs without incurring the huge costs associated with retraining\citep{bourtoule2021machine, cha2024towards,liu2025enabling}. The LLM unlearning problem is illustrated in~\Cref{fig:intro}, which depicts how a model is required to erase a targeted piece of knowledge (e.g., ``Capital of France $\rightarrow$ Paris'') while preserving other unrelated, general knowledge.

Traditional MU methods typically involve complete retraining or extensive fine-tuning of the model, using negative gradients on data points intended for removal~\citep{ginart2019making,graves2021amnesiac}. While these methods are theoretically effective, they become impractical due to high computational complexity, especially as the parameter size of modern LLMs grows exponentially~\citep{yao2024machine}. For instance, exact retraining approaches are computationally prohibitive for billion-scale models due to the enormous data storage and GPU computation required. Recent attempts to mitigate these costs have proposed approximate unlearning methods, including gradient ascent targeting the influence of particular data points~\citep{yao2024machine}, influence function-based approaches leveraging Fisher information matrices~\citep{liu2022right, guo2019certified}, and data subset partitioning strategies such as SISA (Sharded, Isolated, Sliced, and Aggregated)~\citep{bourtoule2021machine}. Despite their improvements, these methods generally depend on full access to the model's gradient information or second-order approximations, which remain computationally burdensome and memory-intensive~\citep{mehta2022deep}.

Driven by these practical challenges, the core problem becomes more fundamental:
\textbf{\uppercase\expandafter{\romannumeral1})}:
\textit{How can we effectively reconstruct full-model gradients from minimal parameter updates for efficient unlearning?}
In typical MU scenarios, directly computing or storing gradients of an LLM with billions of parameters is impractical. Therefore, we seek a mechanism to reconstruct approximate gradients from lightweight, localized model adjustments. \textbf{\uppercase\expandafter{\romannumeral2})}:
\textit{How can we ensure that reconstructed gradients generalize effectively from proxy models to the original large-scale LLMs?}
Gradient approximation often involves training surrogate or proxy models; however, the transferability of gradient signals across model architectures or parameterizations remains an open research question~\citep{wu2020towards, wu2018understanding}.

To address these key questions, we introduce a novel and efficient MU framework named \textit{Recover-to-Forget} ({\ourmethod}). The central insight of {\ourmethod} is leveraging Low-Rank Adaptation (LoRA) modules~\citep{hu2022lora} as a compact representation of gradient updates, dramatically reducing memory and computational costs. Specifically, we exploit the intrinsic low-rank structure of model parameter updates induced by targeted inputs, and then reconstruct the full model gradients using a specialized gradient decoder trained on a smaller-scale proxy model. This decoder maps low-dimensional LoRA representations to high-dimensional gradient approximations, enabling rapid, memory-efficient parameter adjustments during unlearning.

The rationale behind our design is twofold:
(i) Prior studies have demonstrated that low-rank representations effectively capture essential model behaviors, ensuring minimal loss of knowledge while reducing redundancy~\citep{aghajanyan2020intrinsic, zhang2023fine};
(ii) Gradients inherently encode the directional signals required for effective parameter updates, making them ideal candidates for accurate reconstruction and transfer across model scales~\citep{kirkpatrick2017overcoming, liu2022right, yao2024machine}. By combining these insights, our {\ourmethod} framework efficiently performs targeted unlearning without requiring costly full-model gradient computations or retraining. To summarize, our primary contributions include:
\begin{fullitemize}
\item We propose \textit{Recover-to-Forget} ({\ourmethod}), a novel unlearning framework for large-scale LLMs that leverages low-rank adaptations for efficient gradient reconstruction and selective forgetting.
\item We develop a lightweight gradient decoder, trained on a proxy model, capable of reconstructing accurate full-model gradients from compact low-rank parameter representations.
\item We provide theoretical analysis and empirical validation of the gradient transferability from proxy models to large-scale target LLMs, ensuring the broad applicability of our approach.
\item We conduct comprehensive empirical evaluations on established benchmark datasets, demonstrating that {\ourmethod} achieves superior unlearning efficacy and computational efficiency compared to state-of-the-art methods.
\end{fullitemize}

\section{Methodology}
\begin{figure*}[t]
    \includegraphics[trim=30 300 30 20,clip,width=\linewidth]{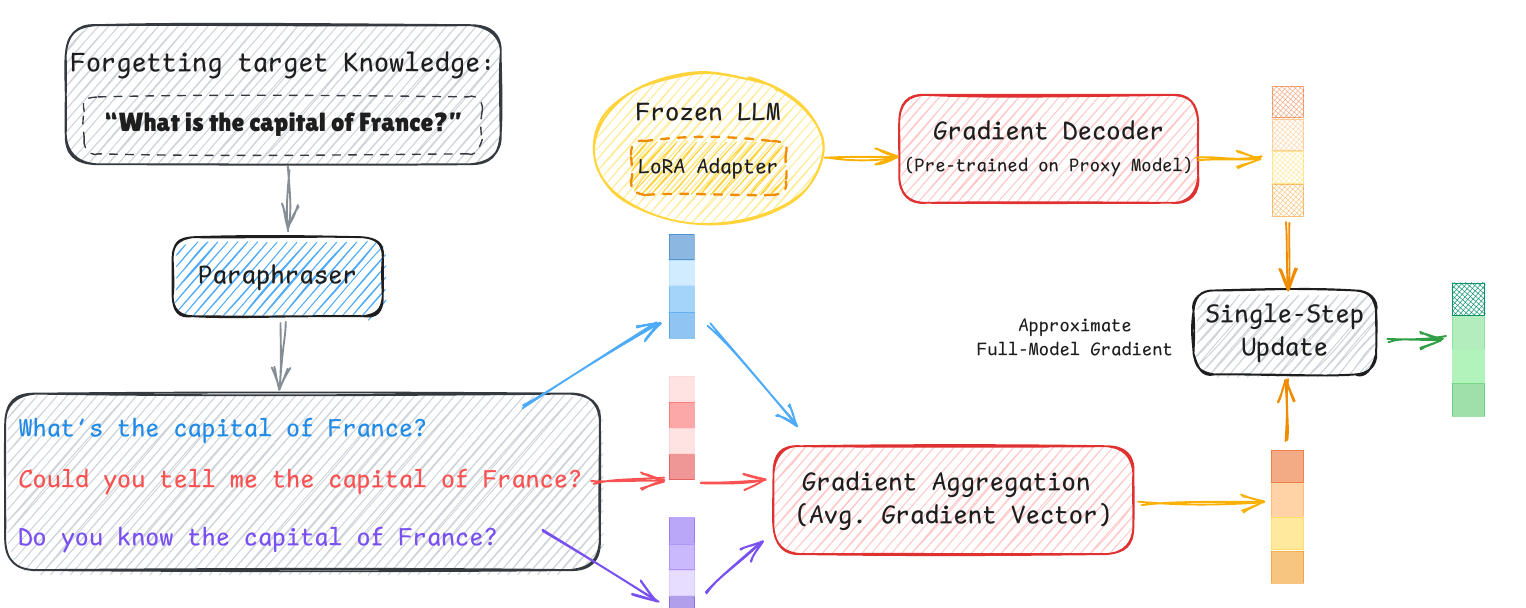}\label{fig:framework}
    \caption{The Recover-to-Forget ({\ourmethod}) framework. Given a target knowledge, {\ourmethod} generates paraphrased queries, extracts LoRA gradients from a frozen LLM, and aggregates them. A Gradient Decoder reconstructs the full-model gradient, which is used to perform a single-step update to forget the target knowledge.}
    \label{fig:r2f}
      \vspace{-1em}
\end{figure*}
We introduce \textbf{Recover-to-Forget ({\ourmethod})}, a novel framework for efficient unlearning in LLMs. As illustrated in~\Cref{fig:r2f}, {\ourmethod} reconstructs full-model gradients from LoRA-based low-rank updates using a proxy model and a gradient decoder, enabling targeted forgetting without modifying the original model weights.

\subsection{Preliminaries}
\noindent\textbf{LLM Training}. Consider a large language model parameterized by $\theta \in \mathbb{R}^d$ that maps inputs $x \in \mathcal{X}$ to outputs $y \in \mathcal{Y}$. Training typically involves optimizing parameters $\theta$ by minimizing a loss function $\mathcal{L}(\theta)$ defined over a training dataset $\mathcal{D}$:
\begin{equation}
\theta^* = \arg\min_{\theta} \mathbb{E}_{(x,y)\sim \mathcal{D}}\left[\mathcal{L}(x, y; \theta)\right].
\end{equation}

\noindent\textbf{Low-Rank Adaptation (LoRA).}
LoRA~\citep{hu2022lora} introduces parameter-efficient fine-tuning by representing the updated weight matrix $W' \in \mathbb{R}^{d \times d}$ as a low-rank perturbation of the original frozen weight $W$. 
Instead of learning a full $d \times d$ update, LoRA injects trainable low-rank modules into selected projection layers (e.g., attention or feed-forward), so that task-specific signals are captured in a compact subspace while the pretrained backbone remains intact. 
This design reduces the number of trainable parameters, lowers memory traffic, and makes it easy to share one base model across many LoRA adapters:
\begin{equation}
    W' = W + AB,
\end{equation}
where $A \in \mathbb{R}^{d \times r}, B \in \mathbb{R}^{r \times d}$ are low-rank matrices with $r \ll d$. During fine-tuning, $W$ is frozen and only $A, B$ are trained, greatly reducing the number of trainable parameters.

\noindent\textbf{Notation.}
Given a pretrained large language model (LLM) with parameters $\theta^* \in \mathbb{R}^d$, we denote by $(x_{\text{tar}}, y_{\text{tar}})$ a target datapoint (or a small target set) whose associated knowledge should be removed.
The model is trained to minimize a task-specific loss $\mathcal{L}(x, y; \theta)$.
We use $|\cdot|$ to denote the Euclidean norm on gradient vectors.
During unlearning, our goal is to estimate (or closely approximate) the full-model gradient
$\nabla_{\theta} \mathcal{L}(x_{\text{tar}}, y_{\text{tar}}; \theta^*)$
\emph{without} performing backpropagation through the full model.

\subsection{Problem Definition: LLM Unlearning}
We define \emph{LLM unlearning} as the process of removing specific knowledge associated with a target datapoint (or a small target set) $(x_{\text{tar}}, y_{\text{tar}})$ from a pretrained model, without retraining the model from scratch and while avoiding degradation on non-target data.
Intuitively, after unlearning, the model should no longer output the original, to-be-forgotten answer for $x_{\text{tar}}$, while behaving similarly to the original model on a \emph{retain set} (i.e., general or unrelated inputs).

Formally, given pretrained parameters $\theta^*$, the unlearning objective is to obtain parameters $\theta_{\text{unlearned}}$ that forget $(x_{\text{tar}}, y_{\text{tar}})$ while preserving overall task accuracy.
A first-order view treats unlearning as taking a small, targeted step along an \emph{unlearning loss}:
\begin{equation}\label{eq:unlearned}
    \theta_{\text{unlearned}} \approx
    \theta^* - \eta \,\nabla_{\theta} \mathcal{L}(x_{\text{tar}}, y_{\text{tar}}; \theta^*),
\end{equation}
where $\eta$ is a step size.%
\footnote{Here $y_{\text{tar}}$ denotes an \emph{unlearning label}, which can be chosen to reduce confidence in the original answer (e.g., a counterfactual label or a softened target), so that gradient descent on $\mathcal{L}(x_{\text{tar}}, y_{\text{tar}};\theta)$ drives the model away from the memorized behavior.}
This formulation views unlearning as applying a targeted update based on the loss incurred by the to-be-forgotten example.
The key difficulty is that, for large LLMs, computing or applying $\nabla_{\theta} \mathcal{L}$ exactly typically requires expensive full-model backpropagation; thus, practical unlearning methods must accurately \emph{approximate} this gradient while still enforcing the preservation constraint on non-target data.

\subsection{Recover-to-Forget: Gradient Reconstruction via Low-Rank Recovery}\label{sec:r2f_proof}
\begin{wrapfigure}[18]{r}{0.45\textwidth}
    \vspace{-2em}
   {\includegraphics[width=0.95\linewidth]{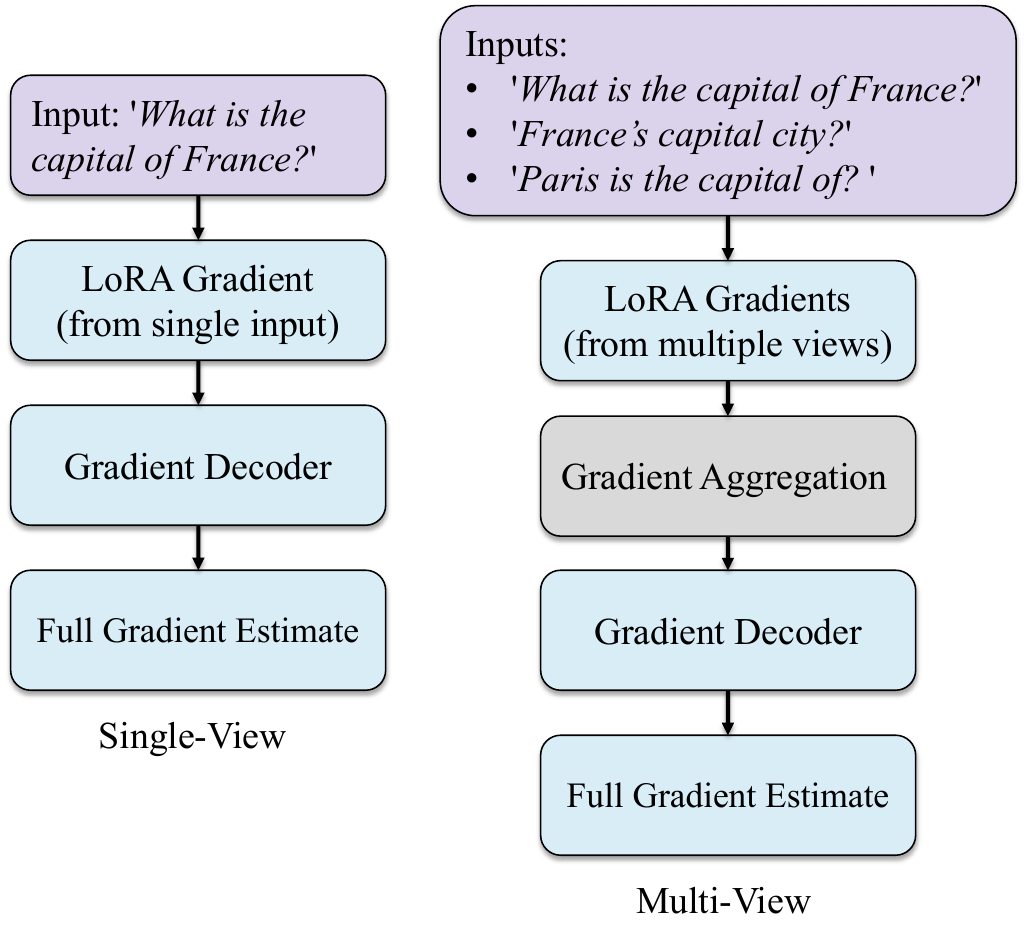}}\label{fig:single_milti}
      \vspace{-1em}
    \caption{Comparison of single-view vs. multi-view gradient reconstruction in Recover-to-Forget. Single-view uses one input for LoRA gradient estimation, while multi-view aggregates gradients from paraphrased inputs, enabling more robust full-gradient recovery.}
    \label{fig:view}
\end{wrapfigure}
We propose \textbf{Recover-to-Forget} (\ourmethod), a parameter-efficient unlearning framework that reconstructs full-model gradients from LoRA updates.
Our method consists of two core steps:
(i) compute low-rank LoRA gradients using multiple paraphrased inputs, and
(ii) train a gradient decoder on a proxy model to reconstruct the corresponding full-model gradient from these low-rank signals.
The overall procedure is summarized in Algorithm~\ref{alg:recover-to-forget}.

\paragraph{Step 1: Multi-view LoRA gradient computation.}
For a target datapoint $(x_{\text{tar}}, y_{\text{tar}})$, we generate $N$ paraphrased prompts $\{x_i\}_{i=1}^N$ to capture diverse views of the target knowledge.
Unlike a single-view gradient, which may reflect only a narrow linguistic context, aggregating gradients over multiple paraphrases yields a more comprehensive and robust signal that better captures the underlying semantic concept~\citep{ilyas2019adversarial, chen2020simple, jaiswal2020survey}.

For example, if the target knowledge is queried as
``What is the capital of France?'', paraphrases such as
``Paris is the capital of which country?'',
``Name the capital city of France.'', and
``France’s capital city is?''
provide different linguistic contexts while encoding the same fact.

For each paraphrase $x_i$, we compute the LoRA gradients with respect to the low-rank parameters $(A,B)$:
\begin{equation}\label{eq:grad_lora}
  \mathcal{G}_{\text{Lo}}(x_i)
  = \nabla_{A,B} \,\mathcal{L}(x_i, y_{\text{tar}}; \theta^*, A, B).
\end{equation}
We then aggregate these gradients by averaging across views:
\begin{equation}
  \bar{\mathcal{G}}_{\text{Lo}}
  = \frac{1}{N} \sum_{i=1}^{N} \mathcal{G}_{\text{Lo}}(x_i).
\end{equation}

\paragraph{Step 2: Gradient decoder via proxy model.}
To reconstruct the corresponding full-model gradient from LoRA gradients, we introduce a gradient decoder network $f_\phi$ parameterized by $\phi$, trained on a smaller proxy model.
The decoder learns a mapping
\begin{equation}
  \hat{\mathcal{G}}_{\text{full}}
  = f_\phi(\bar{\mathcal{G}}_{\text{Lo}})
  \in \mathbb{R}^d,
\end{equation}
where $\hat{\mathcal{G}}_{\text{full}}$ aims to approximate the true full-model gradient
$\nabla_{\theta}\mathcal{L}(x_{\text{tar}}, y_{\text{tar}}; \theta^*)$.

During training, we collect pairs of LoRA and full-model gradients from the proxy model and fit the decoder by minimizing the mean squared error (MSE):
\begin{equation}\label{eq:objective}
  \phi^* = \arg\min_{\phi}
  \,\mathbb{E}_{x\sim \mathcal{D}_{\text{pro}}}
  \big[\,\big|f_\phi(\mathcal{G}_{\text{Lo}}(x))
  - \mathcal{G}_{\text{full}}(x)\big|^2\,\big],
\end{equation}
where $\mathcal{D}_{\text{pro}}$ denotes the proxy training distribution,
$\mathcal{G}_{\text{Lo}}(x)$ is the LoRA gradient on the proxy model, and
$\mathcal{G}_{\text{full}}(x)$ is the corresponding full-model gradient.

\medskip
\noindent\textbf{Proposition 1 (Cross-model gradient transfer).}
Let $\mathcal{D}_{\text{pro}}$ and $\mathcal{D}_{\text{tar}}$ denote the gradient distributions of the proxy and target models, respectively.
Define the proxy and target full-model gradients as
$\mathcal{G}_{\text{full}}^{\text{pro}}(x)$ and
$\mathcal{G}_{\text{full}}^{\text{tar}}(x)$.
Then the gradient reconstruction error on the target model satisfies
\begin{equation}
\begin{split}
\mathbb{E}_{x \sim \mathcal{D}_{\text{tar}}}
  \big[\,\big| f_\phi(\mathcal{G}_{\text{Lo}}(x))
      - \mathcal{G}_{\text{full}}^{\text{tar}}(x) \big|\,\big]
&\leq
\mathbb{E}_{x \sim \mathcal{D}_{\text{pro}}}
  \big[\,\big| f_\phi(\mathcal{G}_{\text{Lo}}(x))
      - \mathcal{G}_{\text{full}}^{\text{pro}}(x) \big|\,\big] \\
&\quad + \operatorname{dis}(\mathcal{D}_{\text{pro}}, \mathcal{D}_{\text{tar}})
   + \mathbb{E}_{x \sim \mathcal{D}_{\text{tar}}}
     \big[\,\big|\mathcal{G}_{\text{full}}^{\text{pro}}(x)
           - \mathcal{G}_{\text{full}}^{\text{tar}}(x)\big|\,\big],
\end{split}
\end{equation}
where $\operatorname{dis}(\mathcal{D}_{\text{pro}}, \mathcal{D}_{\text{tar}})$ is a distribution discrepancy term that upper-bounds the difference between expectations of our gradient loss family under $\mathcal{D}_{\text{pro}}$ and $\mathcal{D}_{\text{tar}}$ (e.g., an $\mathcal{H}\Delta\mathcal{H}$-divergence~\citep{ben2006analysis}).

\begin{algorithm}[t]
\caption{Recover-to-Forget (\ourmethod)}
\label{alg:recover-to-forget}
\begin{algorithmic}[1]
\REQUIRE Pretrained LLM parameters $\theta^*$, proxy dataset $\mathcal{D}_{\text{pro}}$, target input $(x_{\text{tar}}, y_{\text{tar}})$, LoRA rank $r$, number of paraphrases $N$, learning rate $\eta$.
\ENSURE Unlearned model parameters $\theta_{\text{unlearned}}$.
\STATE Initialize LoRA parameters $(A, B)$.
\STATE Generate paraphrased inputs $\{x_i\}_{i=1}^N$ for target $(x_{\text{tar}}, y_{\text{tar}})$.
\FOR{each paraphrased input $x_i$}
    \STATE Compute LoRA gradient $\mathcal{G}_{\text{Lo}}(x_i)$ using Eq.~\eqref{eq:grad_lora}.
\ENDFOR
\STATE Compute averaged LoRA gradient
$\bar{\mathcal{G}}_{\text{Lo}} = \frac{1}{N}\sum_{i=1}^{N}\mathcal{G}_{\text{Lo}}(x_i)$.
\STATE Train gradient decoder $f_{\phi}$ on proxy dataset $\mathcal{D}_{\text{pro}}$ as in Eq.~\eqref{eq:objective}, obtaining $\phi^\star$.
\STATE Reconstruct full gradient
$\hat{\mathcal{G}}_{\text{full}} = f_{\phi^*}(\bar{\mathcal{G}}_{\text{Lo}})$.
\STATE Update parameters to unlearn target knowledge:
$\theta_{\text{unlearned}} = \theta^* - \eta\,\hat{\mathcal{G}}_{\text{full}}$ (cf.~Eq.~\eqref{eq:unlearned}).
\RETURN $\theta_{\text{unlearned}}$.
\end{algorithmic}
\end{algorithm}

\noindent\textbf{Proof of Proposition 1.}
By the triangle inequality, for any $x \sim \mathcal{D}_{\text{tar}}$,
\begin{equation}
\begin{split}
  \big|f_\phi(\mathcal{G}_{\text{Lo}}(x))
      - \mathcal{G}_{\text{full}}^{\text{tar}}(x)\big|
&\leq
  \big|f_\phi(\mathcal{G}_{\text{Lo}}(x))
      - \mathcal{G}_{\text{full}}^{\text{pro}}(x)\big|
+ \big|\mathcal{G}_{\text{full}}^{\text{pro}}(x)
      - \mathcal{G}_{\text{full}}^{\text{tar}}(x)\big|.
\end{split}
\end{equation}
Taking expectation over $x \sim \mathcal{D}_{\text{tar}}$ yields
\begin{equation}
\begin{split}
\mathbb{E}_{x \sim \mathcal{D}_{\text{tar}}}
  \big[\,\big|f_\phi(\mathcal{G}_{\text{Lo}}(x))
      - \mathcal{G}_{\text{full}}^{\text{tar}}(x)\big|\,\big]
&\leq
\mathbb{E}_{x \sim \mathcal{D}_{\text{tar}}}
  \big[\,\big|f_\phi(\mathcal{G}_{\text{Lo}}(x))
      - \mathcal{G}_{\text{full}}^{\text{pro}}(x)\big|\,\big] \\
&\quad +
\mathbb{E}_{x \sim \mathcal{D}_{\text{tar}}}
  \big[\,\big|\mathcal{G}_{\text{full}}^{\text{pro}}(x)
        - \mathcal{G}_{\text{full}}^{\text{tar}}(x)\big|\,\big].
\end{split}
\end{equation}
By the definition of the discrepancy term
$\operatorname{dis}(\mathcal{D}_{\text{pro}}, \mathcal{D}_{\text{tar}})$ over our loss family
$g_\phi(x) = \big|f_\phi(\mathcal{G}_{\text{Lo}}(x))
- \mathcal{G}_{\text{full}}^{\text{pro}}(x)\big|$,
we have
\begin{equation}
\begin{split}
\mathbb{E}_{x \sim \mathcal{D}_{\text{tar}}}
  \big[\,\big|f_\phi(\mathcal{G}_{\text{Lo}}(x))
      - \mathcal{G}_{\text{full}}^{\text{pro}}(x)\big|\,\big]
&\leq
\mathbb{E}_{x \sim \mathcal{D}_{\text{pro}}}
  \big[\,\big|f_\phi(\mathcal{G}_{\text{Lo}}(x))
      - \mathcal{G}_{\text{full}}^{\text{pro}}(x)\big|\,\big] \\
&\quad +
\operatorname{dis}(\mathcal{D}_{\text{pro}}, \mathcal{D}_{\text{tar}}),
\end{split}
\end{equation}
which gives the desired bound after combining the two inequalities.
\hfill$\square$

\subsection{Multi-View Input Generation}
To further improve gradient reconstruction in the \ourmethod{} framework, we adopt a multi-view strategy by generating diverse paraphrases of the target input.

\noindent\textbf{Paraphrase generation.}
We use neural paraphrasing models such as T5 and BART, trained on large-scale datasets like ParaNMT-50M~\citep{wieting2017paranmt}, to produce semantically consistent variants of the target prompt.

\noindent\textbf{Filtering and effectiveness.}
We apply semantic-similarity filtering (e.g., cosine similarity in an embedding space) to ensure paraphrases remain faithful to the original input while providing lexical and syntactic diversity.
These diverse views yield richer LoRA gradients and enable the gradient decoder to better reconstruct full-model gradients, which empirically improves unlearning performance in terms of both removal effectiveness and utility retention.

\section{Experiment}
\vspace{-0.3em}
In this section, we evaluate the proposed {\ourmethod} on four LLM unlearning benchmarks, compare it with representative baselines, and analyze the effect of LoRA rank, multi-view inputs, and gradient reconstruction.

\subsection{Benchmarks and Metrics}
We consider four public and diverse unlearning datasets that cover factual removal, safety/memorization, and privacy-sensitive scenarios: RWKU~\citep{jin2024rwku}, WMDP, MUSE, and WaterDrum. These datasets collectively test whether a method can remove a \emph{target} piece of knowledge while keeping the model usable on general prompts; detailed dataset descriptions are deferred to \Cref{app:datasets}.

To measure both forgetting and utility, we follow standard practice and report four metrics: (i) \textbf{USR} (Unlearning Success Rate) to quantify how often the target answer is no longer produced; (ii) \textbf{GUR} (General Utility Retention) to check that performance on non-target data is preserved; (iii) \textbf{RAP} (Relearning Attack Precision), which re-fine-tunes the unlearned model on a small paraphrased target set and tests whether the removed fact is easily recovered; and (iv) \textbf{MIA} (Model Identity Alignment), which compares the outputs of the original and unlearned models on generic prompts to ensure behavior is not overly distorted. Full metric definitions and protocol details, including the exact RAP and MIA procedures, are provided in \Cref{app:metrics}.

\subsection{Baselines}
We compare {\ourmethod} with both full-gradient and parameter-efficient unlearning approaches. The baselines include (1) a \emph{naive full-gradient} method that backpropagates through the whole LLM on the target example, (2) \emph{single-view LoRA} unlearning that applies LoRA-based updates on one target prompt, and (3) \emph{multi-view LoRA} unlearning that aggregates LoRA gradients over several paraphrased variants to improve robustness. We also include recent inference-time or localized unlearning methods when applicable. This set covers the trade-off between accuracy of forgetting and computational cost; the complete baseline list and hyperparameters are summarized in \Cref{app:baselines}.

\subsection{Experimental Setup}\label{sec:main_setup}
For fair comparison, all methods are run on the same 7B-scale LLM and trained under a unified protocol. Unless otherwise stated, we use LoRA with rank $r=8$ and apply $5$ paraphrased views per target sample to reduce sensitivity to wording. Our gradient decoder is trained using a lightweight \emph{proxy model} (e.g., Mistral-3B) that shares the architecture with the target model but is fully accessible: we collect both LoRA gradients and full gradients on this proxy model over a held-out set of $1$k examples and use them as supervision to learn the mapping from low-rank to full-model gradients. This allows {\ourmethod} to avoid repeated full-model backpropagation on the large target LLM while still producing high-quality reconstructed gradients. All experiments are repeated three times with different seeds, and we report the mean and standard deviation. Training is conducted on NVIDIA RTX A4000 GPUs (16GB). Additional implementation details, dataset-specific batch sizes, and ablation settings are deferred to \Cref{app:exp-setup}.

\begin{table*}[t]
\centering
\setlength{\tabcolsep}{2pt}
\renewcommand{\arraystretch}{1.2}
\scalebox{0.86}{
\begin{tabular}{l||ccIccIccIcc}
\Xhline{1.2pt}
\rowcolor{tablehead!20} &
\multicolumn{2}{cI}{\textbf{RWKU}} &
\multicolumn{2}{cI}{\textbf{WMDP}} &
\multicolumn{2}{cI}{\textbf{MUSE}} &
\multicolumn{2}{c}{\textbf{WaterDrum}} \\
\cline{2-3}\cline{4-5}\cline{6-7}\cline{8-9}
\rowcolor{tablehead!20} \multirow{-2}{*}{\textbf{Method}} 
& {\textbf{USR} $\uparrow$} & {\textbf{GUR} $\uparrow$}
& {\textbf{USR} $\uparrow$} & {\textbf{GUR} $\uparrow$}
& {\textbf{USR} $\uparrow$} & {\textbf{GUR} $\uparrow$}
& {\textbf{USR} $\uparrow$} & {\textbf{GUR} $\uparrow$} \\
\hline\hline
\rowcolor{gray!10}{\full}        
& $84.7 \pm 0.5$ & $91.1 \pm 0.4$
& $82.3 \pm 0.6$ & $90.4 \pm 0.5$
& $79.6 \pm 0.7$ & $89.5 \pm 0.4$
& $83.1 \pm 0.5$ & $91.0 \pm 0.4$ \\
{\loras}       
& $68.2 \pm 0.6$ & $95.3 \pm 0.3$
& $65.7 \pm 0.7$ & $94.7 \pm 0.4$
& $62.0 \pm 0.5$ & $93.1 \pm 0.3$
& $66.5 \pm 0.6$ & $94.0 \pm 0.3$ \\
\rowcolor{gray!10}{\loram}       
& $74.9 \pm 0.5$ & $94.8 \pm 0.4$
& $71.5 \pm 0.6$ & $94.2 \pm 0.4$
& $69.2 \pm 0.6$ & $92.8 \pm 0.4$
& $73.8 \pm 0.5$ & $93.7 \pm 0.3$ \\
{\scrub}       
& $81.3 \pm 0.4$ & $88.6 \pm 0.5$
& $77.1 \pm 0.5$ & $87.2 \pm 0.4$
& $75.9 \pm 0.6$ & $86.8 \pm 0.4$
& $78.0 \pm 0.5$ & $87.9 \pm 0.3$ \\
\rowcolor{gray!10}{\eco}         
& $69.5 \pm 0.6$ & $93.9 \pm 0.3$
& $64.8 \pm 0.7$ & $94.1 \pm 0.4$
& $61.5 \pm 0.5$ & $91.3 \pm 0.3$
& $65.2 \pm 0.6$ & $92.2 \pm 0.3$ \\
{\sku}         
& $76.2 \pm 0.5$ & $92.0 \pm 0.4$
& $73.3 \pm 0.6$ & $91.5 \pm 0.4$
& $70.0 \pm 0.6$ & $90.1 \pm 0.4$
& $74.5 \pm 0.5$ & $90.4 \pm 0.4$ \\
\rowcolor{gray!10}{\cut}         
& $80.1 \pm 0.4$ & $89.4 \pm 0.5$
& $76.6 \pm 0.5$ & $88.3 \pm 0.5$
& $74.2 \pm 0.6$ & $87.9 \pm 0.5$
& $76.3 \pm 0.4$ & $88.6 \pm 0.4$ \\
\rowcolor[HTML]{D7F6FF}{\ourmethod}
& $\mathbf{89.3 \pm 0.3}$ & $\mathbf{95.7 \pm 0.2}$
& $\mathbf{86.5 \pm 0.4}$ & $\mathbf{95.0 \pm 0.3}$
& $\mathbf{84.1 \pm 0.4}$ & $\mathbf{94.6 \pm 0.3}$
& $\mathbf{87.4 \pm 0.3}$ & $\mathbf{95.3 \pm 0.2}$ \\
\Xhline{1.2pt}
\end{tabular}}
\caption{\textbf{Evaluation of unlearning methods on four datasets.} USR ($\uparrow$): unlearning success; GUR ($\uparrow$): general utility retention. Results are mean $\pm$ std over 3 runs. \textbf{Bold} indicates the best score. {\full} denotes Naive Full Gradient; {\loras}/{\loram} denote single-/multi-view LoRA.}
\label{tab:unlearning_results}
\vspace{-1.5em}
\end{table*}

\subsection{Unlearning and Utility Effectiveness}
We analyze the performance of {\ourmethod} in comparison with baseline methods across four datasets: {RWKU}, {WMDP}, {MUSE}, and {WaterDrum}, using USR and GUR. {\ourmethod} outperforms all baselines in unlearning success while maintaining high utility and scalability. While full-gradient methods are effective but costly, and single-view LoRA lacks gradient completeness, {\ourmethod} bridges the gap by reconstructing full gradients from efficient low-rank updates.

{\ourmethod} achieves the highest USR across all datasets, e.g., 86.5\% on {WMDP} vs. 82.3\% (Naive Full Gradient) and 71.5\% (Multi-View LoRA), demonstrating its ability to capture more complete target knowledge through multi-view gradient aggregation. Despite strong unlearning, {\ourmethod} maintains GUR between 94.6\% and 95.7\%, matching or exceeding LoRA-based methods and outperforming full-gradient baselines like {\scrub} and {\cut}, which suffer greater utility drops.

\vspace{-0.3em}
\subsection{Relearning Attack Precision (RAP)}
\begin{wraptable}{r}{0.5\textwidth}
\vspace{-1em}
\centering
\small
\setlength{\tabcolsep}{3pt}
\renewcommand{\arraystretch}{1.0}
\begin{tabular}{l||cccc}
\Xhline{1.2pt}
\rowcolor{tablehead!20}
\textbf{Method} & \textbf{RWKU} & \textbf{WMDP} & \textbf{MUSE} & \textbf{WaterDrum} \\
\hline\hline
\rowcolor{gray!10}{\full}      & $21.4$ & $23.6$ & $25.1$ & $22.9$ \\
{\loras}     & $38.9$ & $41.2$ & $43.0$ & $39.5$ \\
\rowcolor{gray!10}{\loram}     & $30.3$ & $33.7$ & $35.2$ & $31.0$ \\
{\scrub}     & $26.5$ & $28.1$ & $31.6$ & $27.9$ \\
\rowcolor{gray!10}{\eco}       & $36.4$ & $40.2$ & $41.5$ & $38.1$ \\
{\sku}       & $28.0$ & $30.5$ & $33.0$ & $30.0$ \\
\rowcolor{gray!10}{\cut}       & $24.9$ & $26.8$ & $29.2$ & $25.7$ \\
\rowcolor[HTML]{D7F6FF}\textbf{{\ourmethod} (Ours)}
& $\mathbf{18.3}$ & $\mathbf{20.1}$ & $\mathbf{22.5}$ & $\mathbf{19.4}$ \\
\Xhline{1.2pt}
\end{tabular}
\caption{\textbf{Relearning Attack Precision (RAP)} ($\downarrow$), lower is better.}
\label{tab:rap}
\vspace{-1em}
\end{wraptable}
\Cref{tab:rap} shows that \textbf{Recover-to-Forget ({\ourmethod})} achieves the lowest RAP scores across all datasets, indicating the strongest resistance to relearning attacks. For example, on the {MUSE} dataset, {\ourmethod} obtains a RAP of 22.5\%, significantly outperforming Multi-View LoRA (35.2\%) and {\scrub} (31.6\%). This suggests that {\ourmethod} performs a deeper, more irreversible form of knowledge removal. In contrast, lightweight baselines such as {\eco} and Single-View LoRA are more prone to re-injection of forgotten knowledge, as their parameter updates are limited to narrow low-rank subspaces and are easier to reverse through prompting or adversarial attacks.

\vspace{-0.3em}
\subsection{Model Identity Alignment (MIA)}
\begin{wraptable}{r}{0.52\textwidth}
\vspace{-1em}
\centering
\small
\setlength{\tabcolsep}{3pt}
\renewcommand{\arraystretch}{1.0}
\begin{tabular}{l||cccc}
\Xhline{1.2pt}
\rowcolor{tablehead!20}
\textbf{Method} & \textbf{RWKU} & \textbf{WMDP} & \textbf{MUSE} & \textbf{WaterDrum} \\
\hline\hline
\rowcolor{gray!10}{\full}       & $0.091$ & $0.085$ & $0.098$ & $0.089$ \\
{\loras}      & $\mathbf{0.031}$ & $\mathbf{0.029}$ & $\mathbf{0.034}$ & $\mathbf{0.030}$ \\
\rowcolor{gray!10}{\loram}      & $0.045$ & $0.042$ & $0.048$ & $0.043$ \\
{\scrub}      & $0.063$ & $0.061$ & $0.067$ & $0.060$ \\
\rowcolor{gray!10}{\eco}        & $0.036$ & $0.034$ & $0.037$ & $0.033$ \\
{\sku}        & $0.051$ & $0.047$ & $0.054$ & $0.048$ \\
\rowcolor{gray!10}{\cut}        & $0.072$ & $0.068$ & $0.076$ & $0.069$ \\
\rowcolor[HTML]{D7F6FF}{\ourmethod} (Ours)
& $0.053$ & $0.049$ & $0.057$ & $0.051$ \\
\Xhline{1.2pt}
\end{tabular}
\caption{\textbf{Model Identity Alignment (MIA)} ($\downarrow$), lower indicates more change from the original model.}
\label{tab:mia}
\vspace{-1em}
\end{wraptable}
As shown in~\Cref{tab:mia}, {\ourmethod} achieves moderate MIA values across all datasets, indicating a balanced update that removes targeted knowledge without disrupting unrelated behavior. While Single-View LoRA and {\eco} obtain the lowest MIA scores, their corresponding low USR and high RAP suggest they do not adequately forget the targeted information. On the other hand, methods like {\cut} and Naive Full Gradient induce larger shifts, possibly degrading performance on other tasks. {\ourmethod} achieves a desirable trade-off: it alters the model just enough to forget the necessary knowledge while keeping its general behavior intact.

\subsection{Ablation Study}
To understand which design choices matter most in {\ourmethod}, we run ablations on (i) the LoRA rank $r$, (ii) the number of paraphrased views used to estimate the unlearning direction, and (iii) the effect of proxy–target alignment on gradient reconstruction (see \Cref{app:proxy-transfer}). Below, we focus on the first two, since they directly control capacity and semantic coverage.

\subsubsection{Effect of LoRA Rank}
\begin{figure*}[t]
   {\includegraphics[width=1.0\linewidth]{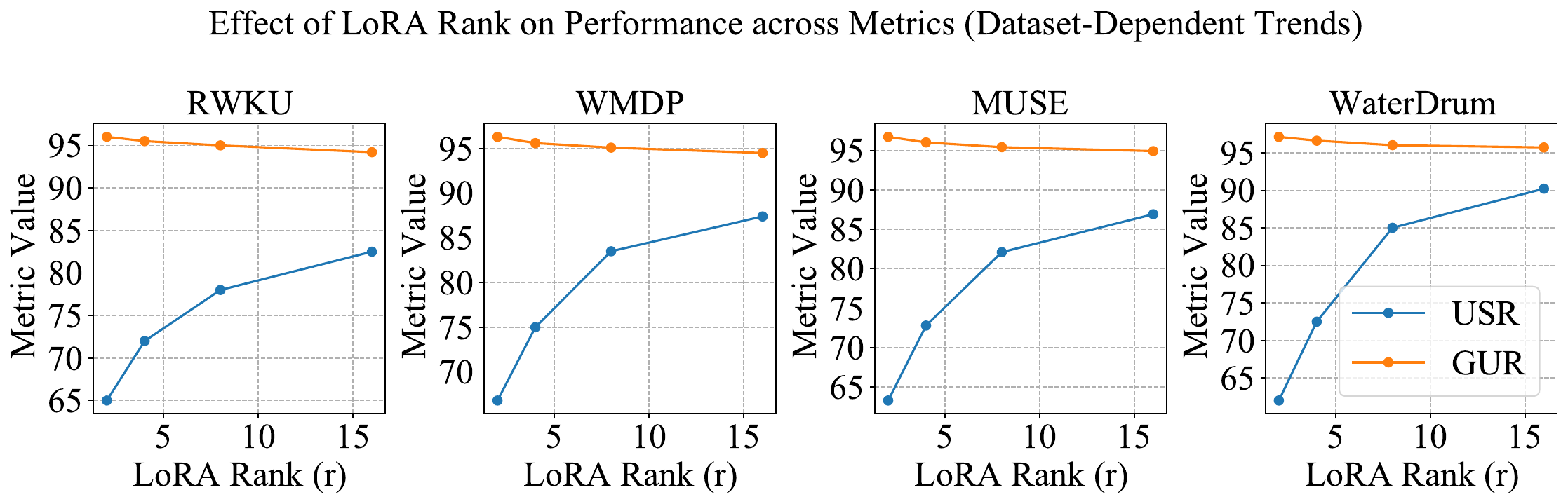}}\label{intro:dp}
    \vspace{-1em}
    \caption{\textbf{Effect of LoRA rank on \ourmethod.}
Each subfigure shows the trends of four evaluation metrics: Unlearning Success Rate (USR), General Utility Retention (GUR), Relearning Attack Precision (RAP), and Model Identity Alignment (MIA), as LoRA rank increases from $2$ to $16$.
Larger datasets (e.g., {MUSE} and {WaterDrum}) demonstrate steeper gains in USR with minimal degradation in GUR, indicating more effective and stable unlearning.
Smaller datasets (e.g., {RWKU}) reveal a sharper trade-off between forgetting and utility retention.}
    \label{fig:lora_rank}
      \vspace{-1em}
\end{figure*}

We vary the LoRA rank from 2 to 16 and re-run R2F on all four datasets (RWKU, WMDP, MUSE, WaterDrum); \Cref{fig:lora_rank} reports USR and GUR. Increasing the rank enlarges the adaptation subspace, so the update can more closely match the full-model unlearning direction.

Overall, the figure shows that LoRA rank mainly controls the forgetting–utility tradeoff. \textbf{(i) More capacity $\Rightarrow$ stronger forgetting}: USR increases on every dataset when $r$ grows, with especially steep gains on MUSE and WaterDrum, meaning higher-rank adapters better capture the target-specific gradient. \textbf{(ii) The utility cost is dataset-dependent}: GUR stays almost flat on large/redundant datasets (MUSE, WaterDrum), but drops a little on smaller RWKU/WMDP, indicating that rich datasets can absorb stronger updates without hurting general behavior. \textbf{(iii) There is a practical sweet spot}: ranks around $r=8$–$12$ already deliver most of the USR gain while keeping GUR within about 1–1.5 points of the base model, so we adopt $r=8$ as the default in the main experiments.

\subsubsection{Effect of Number of Views}
\begin{figure*}[t]
   {\includegraphics[width=1.0\linewidth]{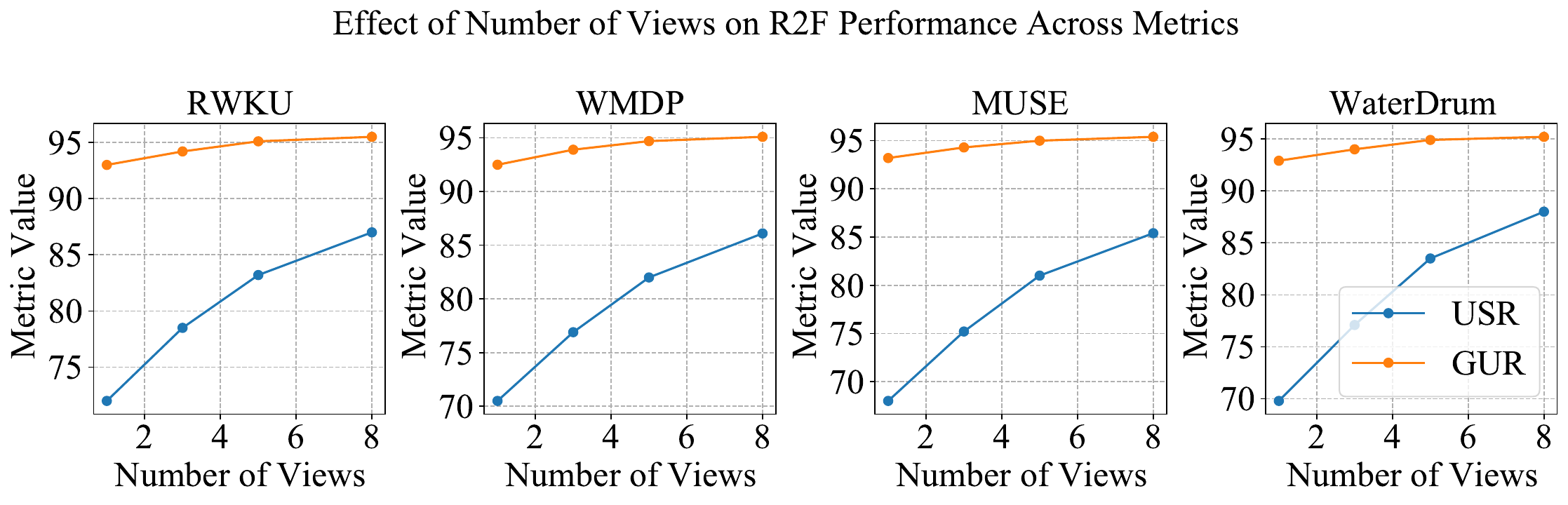}}
      \vspace{-1em}
\caption{%
\textbf{Effect of paraphrased view count on {\ourmethod}.} Each subfigure shows USR, GUR, RAP, and MIA as the number of views increases from 1 to 8. More views consistently improve USR and help preserve GUR. Larger datasets (e.g., {MUSE}, {WaterDrum}) show greater stability, while smaller ones (e.g., {RWKU}) see larger relative gains.
}

    \label{fig:view_results}
        \vspace{-1em}
\end{figure*}

We study how the number of paraphrased target views affects unlearning by increasing the views from 1 to 8 and re-running R2F on all four datasets (RWKU, WMDP, MUSE, WaterDrum); results are shown in \Cref{fig:view_results}. Each view is a semantically equivalent reformulation of the same fact, so more views give the gradient decoder a richer approximation of the target knowledge.

Overall, \Cref{fig:view_results} reveals that adding views makes unlearning both stronger and more stable. \textbf{(i) Multi-view consistently helps forgetting}: USR rises almost monotonically on all datasets; the gain is most obvious on RWKU and WMDP, where single-view LoRA is sensitive to phrasing, but 5--8 views let the decoder see the whole semantic neighborhood. \textbf{(ii) Utility remains high or slightly improves}: GUR lines stay flat (and sometimes improve), indicating that multi-view gradients make the update more targeted, which forgets the intended fact while touching less unrelated behavior. \textbf{(iii) Larger datasets show smaller marginal gains}: on MUSE and WaterDrum, the trend is still positive but smoother, because these datasets already contain diverse contexts, so extra paraphrases provide refinement rather than rescue.

\subsection{Resource-Based Comparison}
\begin{table*}[t]
\centering
\small
\setlength{\tabcolsep}{4pt}
\renewcommand{\arraystretch}{1.0}
\begin{tabular}{l||cccccccc}
\Xhline{1.2pt}
\rowcolor{tablehead!20}
\textbf{Method} & {\full} & {\loras} & {\loram} & {\scrub} & {\eco} & {\sku} & {\cut} & {\ourmethod} \\
\hline\hline
\rowcolor{gray!10}\textbf{Time (s)}
& $5.8$ 
& $\mathbf{2.3}$ 
& $\underline{2.5}$ 
& $3.2$ 
& $4.1$ 
& $3.5$ 
& $3.7$ 
& $2.9$ \\
\textbf{Memory (MB)}
& $14500$
& $\mathbf{1800}$
& $\underline{1900}$
& $2500$
& $3100$
& $2700$
& $3200$
& $2200$ \\
\Xhline{1.2pt}
\end{tabular}
\caption{\textbf{Compute time and memory usage} on RWKU (per-deletion averages). Our method \textbf{\ourmethod} includes gradient decoding and LoRA update; decoder training is a one-time cost. Lower is better.}
\label{tab:rwku_compute}
\end{table*}
We also provide a summary of the compute time and memory usage for R2F and all baselines under the RWKU dataset setting. To ensure a fair comparison, we report the average per-deletion runtime and memory cost, assuming one deletion request per evaluation. For R2F, the reported runtime includes the cost of gradient decoding and LoRA update, which are the only per-deletion steps required. The one-time decoder training (based on proxy model gradients) is lightweight and shared across deletion requests, and thus not included in the per-sample cost.

The~\Cref{tab:rwku_compute} demonstrates that R2F achieves a favorable trade-off: it is significantly more efficient than full-model retraining methods like FullGrad or DeepCUT, while maintaining higher unlearning quality than inference-only methods such as SCRUB and ECO.

\section{Related Work}
We group related efforts into (i) machine unlearning for large language models and (ii) low-rank adaptation with gradient/parameter reconstruction.

\noindent\textbf{Machine unlearning in LLMs.}
Early LLM-focused unlearning works formulated forgetting as pushing the model away from a specific example or fact, typically via gradient ascent or relabeling on the target instance~\citep{eldan2023whos,chen2023unlearn,jang2022knowledge}. 
Subsequent papers pointed out that naïve ascent can cause over-forgetting and that evaluation must separately report forgetting and utility~\citep{liu2024rethinking,maini2024tofu,blanco2025unlearning}. 
A line of methods makes the forgetting step lighter or more controllable: offset-based unlearning edits the logits around the target answer~\citep{huang2024offset,ji2024reversing}, while inference-/prompt-time approaches such as ECO-style operations suppress undesired outputs without touching base weights, at the cost of weaker guarantees~\citep{feng2024economic}. 
Very recent work explores \emph{parameter-efficient} LLM unlearning, i.e., carrying out the forgetting update in a PEFT/LoRA space so that the base model stays frozen and multiple “forget adapters’’ can be swapped in~\citep{ding2025unified,cha2025towards}. 
There is also growing evidence that model editing techniques (e.g., ROME, MEMIT) can be repurposed to remove facts, but they require extra constraints to avoid over-editing and to preserve unrelated behavior~\citep{meng2022locating,meng2022mass}. 
Our work follows this LLM-specific line, but targets the case where we want full-model–quality updates while only observing low-rank / PEFT gradients.

\noindent\textbf{Low-rank adaptation and gradient reconstruction.}
As LLMs continue to scale, improving their efficiency—in terms of compute, memory, and serving cost—has become critical for practical deployment, motivating methods that reduce training and inference overhead without sacrificing performance. 
LoRA enables parameter-efficient fine-tuning by injecting low-rank adapters into attention/MLP projections and training only the small matrices~\citep{hu2022lora}.
Because LoRA lives in a low-dimensional subspace, several recent papers study how to make low-rank updates more expressive or more stable (e.g., PiSSA~\citep{meng2024pissa} and PEFT variants) and how to use them for unlearning~\citep{ding2025unified,cha2025towards}. 
In parallel, the gradient-inversion literature has shown that gradients of deep models, including LLMs, carry enough information to reconstruct inputs~\citep{zhu2019deep,petrov2024dager,xie2025recit}; this naturally suggests using a smaller, fully accessible proxy model to \emph{learn} a mapping from cheap / low-rank gradients to the corresponding full-model direction. 
Proxy-based gradient estimation and gradient matching have been explored earlier in distillation and federated settings, and provide a scalable way to supervise reconstruction for a large, frozen model. 
Our approach is closest in spirit to these proxy/PEFT-based unlearning methods, but we explicitly reconstruct the missing full gradient from LoRA updates so that unlearning can be done with LLM-level fidelity while avoiding repeated full-model backpropagation.
\section{Conclusion}\label{sec:impact}
\vspace{-0.3em}
We proposed \textit{Recover-to-Forget (\ourmethod{})}, a parameter-efficient unlearning framework that reconstructs full-model gradients from LoRA updates using a decoder trained on a proxy model. \ourmethod{} avoids full-model retraining while enabling effective and scalable unlearning in LLMs. Through extensive experiments, we show that \ourmethod{} outperforms full-gradient and lightweight baselines in both removal effectiveness and utility preservation. While \ourmethod{} eliminates the need to access or update full model parameters, its performance relies on the alignment between the proxy and target models.

\section*{Acknowledgements}
This work was supported in part by the DARPA Young Faculty Award, the National Science Foundation (NSF) under Grants \#2127780, \#2319198, \#2321840, \#2312517, and \#2235472, \#2431561, the Semiconductor Research Corporation (SRC), the Office of Naval Research through the Young Investigator Program Award, and Grants \#N00014-21-1-2225 and \#N00014-22-1-2067, Army Research Office Grant \#W911NF2410360. Additionally, support was provided by the Air Force Office of Scientific Research under Award \#FA9550-22-1-0253, along with generous gifts from Xilinx and Cisco.

\newpage
\bibliographystyle{plain}
\bibliography{ref}

\begin{thebibliography}{10}

\bibitem{aghajanyan2020intrinsic}
Armen Aghajanyan, Luke Zettlemoyer, and Sonal Gupta.
\newblock Intrinsic dimensionality explains the effectiveness of language model fine-tuning.
\newblock {\em arXiv preprint arXiv:2012.13255}, 2020.

\bibitem{baumhauer2022machine}
Thomas Baumhauer, Pascal Sch{\"o}ttle, and Matthias Zeppelzauer.
\newblock Machine unlearning: Linear filtration for logit-based classifiers.
\newblock {\em Machine Learning}, 111(9):3203--3226, 2022.

\bibitem{ben2006analysis}
Shai Ben-David, John Blitzer, Koby Crammer, and Fernando Pereira.
\newblock Analysis of representations for domain adaptation.
\newblock {\em NeurIPS}, 19, 2006.

\bibitem{blanco2025unlearning}
Alberto Blanco-Justicia et~al.
\newblock Unlearning in large language models: We are not there yet.
\newblock {\em IEEE Computer}, 2025.

\bibitem{bourtoule2021machine}
Lucas Bourtoule, Varun Chandrasekaran, Christopher~A Choquette-Choo, Hengrui Jia, Adelin Travers, Baiwu Zhang, David Lie, and Nicolas Papernot.
\newblock Machine unlearning.
\newblock In {\em 2021 IEEE symposium on security and privacy (SP)}, pages 141--159. IEEE, 2021.

\bibitem{carlini2021extracting}
Nicholas Carlini, Florian Tramer, Eric Wallace, Matthew Jagielski, Ariel Herbert-Voss, Katherine Lee, Adam Roberts, Tom Brown, Dawn Song, Ulfar Erlingsson, et~al.
\newblock Extracting training data from large language models.
\newblock In {\em 30th USENIX security symposium (USENIX Security 21)}, pages 2633--2650, 2021.

\bibitem{cha2025towards}
Minho Cha, Jihwan Park, Seung~won Kim, and Juho Lee.
\newblock Towards robust and parameter-efficient knowledge unlearning for large language models.
\newblock {\em arXiv preprint arXiv:2502.04112}, 2025.

\bibitem{cha2024towards}
Sungmin Cha, Sungjun Cho, Dasol Hwang, and Moontae Lee.
\newblock Towards robust and cost-efficient knowledge unlearning for large language models.
\newblock {\em arXiv preprint arXiv:2408.06621}, 2024.

\bibitem{chen2023unlearn}
Jiaao Chen and Diyi Yang.
\newblock Unlearn what you want to forget: Efficient unlearning for llms.
\newblock {\em arXiv preprint arXiv:2310.20150}, 2023.

\bibitem{chen2020simple}
Ting Chen, Simon Kornblith, Mohammad Norouzi, and Geoffrey Hinton.
\newblock A simple framework for contrastive learning of visual representations.
\newblock In {\em International conference on machine learning}, pages 1597--1607. PmLR, 2020.

\bibitem{ding2025unified}
Yuxuan Ding, Jiayi Sun, Xian Li, Yicheng Wang, and Zhipeng Chen.
\newblock Unified parameter-efficient unlearning for large language models.
\newblock {\em arXiv preprint arXiv:2503.01854}, 2025.

\bibitem{eldan2023whos}
Ronen Eldan and Mark Russinovich.
\newblock Who's harry potter? approximate unlearning in llms.
\newblock {\em arXiv preprint arXiv:2310.02238}, 2023.

\bibitem{feng2024economic}
Shangyu Feng, Hao Chen, Shiliang Tan, Jiahui Xu, and Bo~Wang.
\newblock Economic corrective operations for llm unlearning.
\newblock {\em arXiv preprint arXiv:2409.01492}, 2024.

\bibitem{ginart2019making}
Antonio Ginart, Melody Guan, Gregory Valiant, and James~Y Zou.
\newblock Making ai forget you: Data deletion in machine learning.
\newblock {\em NeurIPS}, 32, 2019.

\bibitem{gemini3pro2025}
{Google DeepMind}.
\newblock Model evaluation -- approach, methodology \& results: Gemini 3 pro.
\newblock \url{https://deepmind.google/models/evals-methodology/gemini-3-pro}, November 2025.
\newblock Model evaluation report.

\bibitem{graves2021amnesiac}
Laura Graves, Vineel Nagisetty, and Vijay Ganesh.
\newblock Amnesiac machine learning.
\newblock In {\em AAAI}, volume~35, pages 11516--11524, 2021.

\bibitem{guo2019certified}
Chuan Guo, Tom Goldstein, Awni Hannun, and Laurens Van Der~Maaten.
\newblock Certified data removal from machine learning models.
\newblock {\em arXiv preprint arXiv:1911.03030}, 2019.

\bibitem{he2025deep}
Estrid He, Tabinda Sarwar, Ibrahim Khalil, Xun Yi, and Ke~Wang.
\newblock Deep contrastive unlearning for language models.
\newblock {\em arXiv preprint arXiv:2503.14900}, 2025.

\bibitem{hu2022lora}
Edward~J Hu, Yelong Shen, Phillip Wallis, Zeyuan Allen-Zhu, Yuanzhi Li, Shean Wang, Lu~Wang, Weizhu Chen, et~al.
\newblock Lora: Low-rank adaptation of large language models.
\newblock {\em ICLR}, 1(2):3, 2022.

\bibitem{huang2024offset}
James~Y Huang, Wenxuan Zhou, Fei Wang, Fred Morstatter, Sheng Zhang, Hoifung Poon, and Muhao Chen.
\newblock Offset unlearning for large language models.
\newblock {\em arXiv preprint arXiv:2404.11045}, 2024.

\bibitem{ilyas2019adversarial}
Andrew Ilyas, Shibani Santurkar, Dimitris Tsipras, Logan Engstrom, Brandon Tran, and Aleksander Madry.
\newblock Adversarial examples are not bugs, they are features.
\newblock {\em NeurIPS}, 32, 2019.

\bibitem{jagielski2022measuring}
Matthew Jagielski, Om~Thakkar, Florian Tramer, Daphne Ippolito, Katherine Lee, Nicholas Carlini, Eric Wallace, Shuang Song, Abhradeep Thakurta, Nicolas Papernot, et~al.
\newblock Measuring forgetting of memorized training examples.
\newblock {\em arXiv preprint arXiv:2207.00099}, 2022.

\bibitem{jaiswal2020survey}
Ashish Jaiswal, Ashwin~Ramesh Babu, Mohammad~Zaki Zadeh, Debapriya Banerjee, and Fillia Makedon.
\newblock A survey on contrastive self-supervised learning.
\newblock {\em Technologies}, 9(1):2, 2020.

\bibitem{jang2022knowledge}
Joel Jang, Dongkeun Yoon, Sohee Yang, Sungmin Cha, Moontae Lee, Lajanugen Logeswaran, and Minjoon Seo.
\newblock Knowledge unlearning for mitigating privacy risks in language models.
\newblock {\em arXiv preprint arXiv:2210.01504}, 2022.

\bibitem{ji2024reversing}
Yiming Ji, Yuncheng Li, Weizhe Zhang, Jiahui Lu, and Soumen Chakrabarti.
\newblock Reversing the forget--retain objectives: An efficient llm unlearning framework from logit difference.
\newblock In {\em NeurIPS}, 2024.

\bibitem{jin2024rwku}
Zhuoran Jin, Pengfei Cao, Chenhao Wang, Zhitao He, Hongbang Yuan, Jiachun Li, Yubo Chen, Kang Liu, and Jun Zhao.
\newblock Rwku: Benchmarking real-world knowledge unlearning for large language models.
\newblock {\em arXiv preprint arXiv:2406.10890}, 2024.

\bibitem{kirkpatrick2017overcoming}
James Kirkpatrick, Razvan Pascanu, Neil Rabinowitz, Joel Veness, Guillaume Desjardins, Andrei~A Rusu, Kieran Milan, John Quan, Tiago Ramalho, Agnieszka Grabska-Barwinska, et~al.
\newblock Overcoming catastrophic forgetting in neural networks.
\newblock {\em Proceedings of the national academy of sciences}, 114(13):3521--3526, 2017.

\bibitem{laurelli2024brain}
Michele Laurelli.
\newblock Brain surgery: Ensuring gdpr compliance in large language models via concept erasure.
\newblock {\em arXiv preprint arXiv:2409.14603}, 2024.

\bibitem{li2024wmdp}
Nathaniel Li, Alexander Pan, Anjali Gopal, Summer Yue, Daniel Berrios, Alice Gatti, Justin~D Li, Ann-Kathrin Dombrowski, Shashwat Goel, Long Phan, et~al.
\newblock The wmdp benchmark: Measuring and reducing malicious use with unlearning.
\newblock {\em arXiv preprint arXiv:2403.03218}, 2024.

\bibitem{liu2024rethinking}
Siyuan Liu, Sida Zhang, Yang Liu, Liyuan Fan, Jiayi Gu, Yue Sheng, Zexuan Zhong, and Bo~Li.
\newblock Rethinking machine unlearning for large language models.
\newblock {\em arXiv preprint arXiv:2402.08787}, 2024.

\bibitem{liu2023fairgraph}
Yezi Liu.
\newblock {Fairgraph: Automated Graph Debiasing with Gradient Matching}.
\newblock In {\em Proceedings of the 32nd ACM International Conference on Information and Knowledge Management}, pages 4135--4139, 2023.

\bibitem{liu2025lune}
Yezi Liu, Hanning Chen, Wenjun Huang, Yang Ni, and Mohsen Imani.
\newblock {LUNE: Efficient LLM Unlearning via LoRA Fine-Tuning with Negative Examples}.
\newblock In {\em Socially Responsible and Trustworthy Foundation Models at NeurIPS 2025}.

\bibitem{liucauchy}
Yezi Liu, Hanning Chen, Wenjun Huang, Yang Ni, and Mohsen Imani.
\newblock {Cauchy-Schwarz Fairness Regularizer}.
\newblock 2025.

\bibitem{liu2024promoting}
Yezi Liu, Hanning Chen, and Mohsen Imani.
\newblock {Promoting Fairness in Link Prediction with Graph Enhancement}.
\newblock {\em Frontiers in Big Data}, 7:1489306, 2024.

\bibitem{liu2025white}
Yezi Liu, Wenjun Huang, Yang Ni, Hanning Chen, and Mohsen Imani.
\newblock {White admitted by stanford, black got rejections: Exploring racial stereotypes in text-to-image generation from a college admissions lens}.
\newblock In {\em Companion Proceedings of the ACM on Web Conference 2025}, pages 1138--1142, 2025.

\bibitem{liu2025fgu}
Yezi Liu, Prathyush Poduval, Wenjun Huang, Yang Ni, Hanning Chen, and Mohsen Imani.
\newblock {Enabling Group Fairness in Graph Unlearning via Bi-level Debiasing}.
\newblock {\em arXiv preprint arXiv:2505.09702}, 2025.

\bibitem{liu2025enabling}
Yezi Liu and Yanning Shen.
\newblock {Enabling Group Fairness in Machine Unlearning via Distribution Correction}.
\newblock In {\em CIKM}, pages 1925--1935, 2025.

\bibitem{liu2022right}
Yi~Liu, Lei Xu, Xingliang Yuan, Cong Wang, and Bo~Li.
\newblock The right to be forgotten in federated learning: An efficient realization with rapid retraining.
\newblock In {\em IEEE INFOCOM 2022-IEEE conference on computer communications}, pages 1749--1758. IEEE, 2022.

\bibitem{lu2025waterdrum}
Xinyang Lu, Xinyuan Niu, Gregory Kang~Ruey Lau, Bui Thi~Cam Nhung, Rachael Hwee~Ling Sim, Fanyu Wen, Chuan-Sheng Foo, See-Kiong Ng, and Bryan Kian~Hsiang Low.
\newblock Waterdrum: Watermarking for data-centric unlearning metric.
\newblock {\em arXiv preprint arXiv:2505.05064}, 2025.

\bibitem{maini2024tofu}
Pratyush Maini, Zhili Feng, Avi Schwarzschild, Zachary~C Lipton, and J~Zico Kolter.
\newblock Tofu: A task of fictitious unlearning for llms.
\newblock {\em arXiv preprint arXiv:2401.06121}, 2024.

\bibitem{mehta2022deep}
Ronak Mehta, Sourav Pal, Vikas Singh, and Sathya~N Ravi.
\newblock Deep unlearning via randomized conditionally independent hessians.
\newblock In {\em Proceedings of the IEEE/CVF Conference on Computer Vision and Pattern Recognition}, pages 10422--10431, 2022.

\bibitem{meng2024pissa}
Fanxu Meng, Zhaohui Wang, and Muhan Zhang.
\newblock Pissa: Principal singular values and singular vectors adaptation of large language models.
\newblock {\em Advances in Neural Information Processing Systems}, 37:121038--121072, 2024.

\bibitem{meng2022locating}
Kevin Meng, David Bau, Alex Andonian, and Yonatan Belinkov.
\newblock Locating and editing factual associations in gpt.
\newblock {\em NeurIPS}, 35:17359--17372, 2022.

\bibitem{meng2022mass}
Kevin Meng, Arnab~Sen Sharma, Alex Andonian, Yonatan Belinkov, and David Bau.
\newblock Mass-editing memory in a transformer.
\newblock {\em arXiv preprint arXiv:2210.07229}, 2022.

\bibitem{meta2025llama4}
{Meta AI}.
\newblock Llama 4: Multimodal intelligence.
\newblock \url{https://ai.meta.com/blog/llama-4-multimodal-intelligence/}, 2025.
\newblock Blog post.

\bibitem{openai2025gpt51thinking}
{OpenAI}.
\newblock Gpt-5.1 instant and gpt-5.1 thinking system card addendum.
\newblock \url{https://openai.com/index/gpt-5-system-card-addendum-gpt-5-1/}, November 2025.
\newblock System card.

\bibitem{petrov2024dager}
Ivo Petrov, Dimitar~I. Dimitrov, Maximilian Baader, Mark~Niklas M{\"u}ller, and Martin Vechev.
\newblock {DAGER}: Exact gradient inversion for large language models.
\newblock {\em arXiv preprint arXiv:2405.15586}, 2024.

\bibitem{shi2024muse}
Weijia Shi, Jaechan Lee, Yangsibo Huang, Sadhika Malladi, Jieyu Zhao, Ari Holtzman, Daogao Liu, Luke Zettlemoyer, Noah~A Smith, and Chiyuan Zhang.
\newblock Muse: Machine unlearning six-way evaluation for language models.
\newblock {\em arXiv preprint arXiv:2407.06460}, 2024.

\bibitem{wieting2017paranmt}
John Wieting and Kevin Gimpel.
\newblock Paranmt-50m: Pushing the limits of paraphrastic sentence embeddings with millions of machine translations.
\newblock {\em arXiv preprint arXiv:1711.05732}, 2017.

\bibitem{wu2020towards}
Lei Wu and Zhanxing Zhu.
\newblock Towards understanding and improving the transferability of adversarial examples in deep neural networks.
\newblock In {\em Asian Conference on Machine Learning}, pages 837--850. PMLR, 2020.

\bibitem{wu2018understanding}
Lei Wu, Zhanxing Zhu, Cheng Tai, et~al.
\newblock Understanding and enhancing the transferability of adversarial examples.
\newblock {\em arXiv preprint arXiv:1802.09707}, 2018.

\bibitem{xie2025recit}
Jin Xie, Ruishi He, Songze Li, Xiaojun Jia, and Shouling Ji.
\newblock {ReCIT}: Reconstructing full private data from gradient in parameter-efficient fine-tuning of large language models.
\newblock {\em arXiv preprint arXiv:2504.20570}, 2025.

\bibitem{yao2024machine}
Jin Yao, Eli Chien, Minxin Du, Xinyao Niu, Tianhao Wang, Zezhou Cheng, and Xiang Yue.
\newblock Machine unlearning of pre-trained large language models.
\newblock {\em arXiv preprint arXiv:2402.15159}, 2024.

\bibitem{zhang2023fine}
Zhong Zhang, Bang Liu, and Junming Shao.
\newblock Fine-tuning happens in tiny subspaces: Exploring intrinsic task-specific subspaces of pre-trained language models.
\newblock {\em arXiv preprint arXiv:2305.17446}, 2023.

\bibitem{zhu2019deep}
Ligeng Zhu, Zhijian Liu, and Song Han.
\newblock Deep leakage from gradients.
\newblock {\em NeurIPS}, 32, 2019.

\end{thebibliography}

\newpage
\appendix
\section{Datasets and Unlearning Scenarios}
\label{app:datasets}
We evaluate {\ourmethod} on four benchmark LLM unlearning datasets that jointly cover factual removal, safety-critical memorization, and privacy-/identity-sensitive content. 
\begin{fullitemize}

\item  \textbf{RWKU}~\footnote{\url{https://huggingface.co/datasets/jinzhuoran/RWKU}}~\citep{jin2024rwku}
targets the removal of real-world factual associations from LLMs, such as named-entity attributes or factual triples (e.g., ``X was born in Y''). It provides (i) carefully annotated tuples for deletion and (ii) semantically similar control samples for generalization evaluation.

\item \textbf{WMDP}~\footnote{\url{https://github.com/centerforaisafety/wmdp}}~\citep{li2024wmdp} is constructed to test unlearning in safety/memorization settings. It contains highly memorized sentence pairs from Wikipedia that could pose risks if leaked. The task is to forget the risky response while retaining general language understanding.

\item \textbf{MUSE}~\footnote{\url{https://muse-bench.github.io/}}~\citep{shi2024muse}
emphasizes privacy-preserving unlearning by simulating the removal of sensitive personal information. Prompts may include user identity, contact information, or private activities. The goal is to erase these pieces of information while preserving normal response behavior.

\item \textbf{WaterDrum}~\footnote{\url{https://huggingface.co/datasets/Glow-AI/WaterDrum-Ax}}~\citep{lu2025waterdrum} is a large-scale synthetic benchmark for fine-grained, fact-level unlearning. It offers structured Q/A pairs with controlled variations, making it suitable for analyzing the scalability and stability of unlearning methods under different knowledge densities and task structures.

\end{fullitemize}

Collectively, these datasets test whether a method can remove \emph{targeted} knowledge while maintaining model utility across factual, sensitive, and synthetic domains.

\section{Baselines}
\label{app:baselines}
To fairly assess {\ourmethod}, we compare it with representative LLM unlearning methods that span full-gradient, LoRA-based, and inference-/local-edit families.

\paragraph{Naive Full Gradient (Single-View Backprop).}
This baseline backpropagates through the entire pretrained model on a single target instance $(x_{\text{tar}}, y_{\text{tar}})$ and applies the update
$\theta_{\text{unlearned}} = \theta^* - \eta \nabla_\theta \mathcal{L}(x_{\text{tar}}, y_{\text{tar}}; \theta^*)$.
It captures the exact gradient but requires full-model access, high memory, and is sensitive to phrasing, so it does not generalize well.

\paragraph{Single-View LoRA Gradient.}
This method computes the unlearning direction only on LoRA-injected parameters for one target prompt, keeping $\theta^*$ frozen. It is parameter-efficient but only captures a low-rank approximation and may lead to incomplete forgetting when the knowledge is expressed in multiple linguistic forms.

\paragraph{Multi-View LoRA Gradient.}
To improve robustness, this baseline averages LoRA gradients over $N$ paraphrased variants:
\[
\bar{\mathcal{G}}_{\text{Lo}} = \frac{1}{N} \sum_{i=1}^{N} \mathcal{G}_{\text{Lo}}(x_i).
\]
It reduces sensitivity to any single wording but still lives in the LoRA subspace and cannot reconstruct full-model gradients.

\paragraph{SCRUB}\citep{baumhauer2022machine}.
Applies lightweight linear transforms on logits to remove class-specific info without full retraining; requires access to full logits, which is less convenient at LLM scale.

\paragraph{ECO (Embedding-Corrupted Prompts)}\citep{laurelli2024brain}.
An inference-time approach that perturbs prompt embeddings to suppress undesired outputs. It is cheap but may fail when the knowledge is deeply entangled.

\paragraph{SKU (Selective Knowledge Unlearning)}\citep{yao2024machine}.
Edits localized internal representations based on attribution maps to improve precision, but the quality depends on attribution.

\paragraph{DeepCUT}\citep{he2025deep}.
A recent contrastive unlearning method that performs full-model updates with contrastive pairs; effective but computationally heavier.

\section{Evaluation Metrics and Protocols}
\label{app:metrics}
This section expands the brief metric description in the main text and specifies the exact protocols we use.

\begin{fullitemize}
\item \textbf{Unlearning Success Rate (USR).}
Percentage of target prompts for which the model no longer produces the original, to-be-forgotten answer.

\item \textbf{General Utility Retention (GUR).}
Accuracy/quality on a held-out, non-target task to ensure the model remains useful.

\item \textbf{Relearning Attack Precision (RAP).}
We simulate a relearning attack by fine-tuning the \emph{unlearned} model on a small paraphrased target set and then querying the original target. A lower RAP means the forgotten knowledge is harder to reintroduce. We follow the same paraphrasing protocol as in the main experiments; hyperparameters are listed in \Cref{app:exp-setup}.

\item \textbf{Model Identity Alignment (MIA).}
We compute cosine similarity (or a distributional distance) between outputs of $\theta^*$ and $\theta_{\text{unlearned}}$ on 500 general-purpose prompts. A higher MIA means the model behavior stays closer to the original.

\item \textbf{Gradient Ascent Unlearning.}
For completeness, we also report the classical gradient-ascent-style forgetting \citep{jang2022knowledge}, which directly increases the loss on target data but can hurt GUR; we treat it as a reference procedure rather than a main baseline.

\end{fullitemize}

\section{Additional Experimental Setup}\label{app:exp-setup}
\paragraph{Model Specification.}
All experiments use \textbf{Mistral-7B} as the target LLM and a structurally compatible \textbf{3B-scale Mistral} as the proxy. The proxy is \emph{only} used to collect pairs of (LoRA gradient, full gradient) for training the gradient decoder, so it does not participate in inference.

\paragraph{Training Protocol.}
Unless otherwise noted, we set LoRA rank to $r=8$ and apply $5$ paraphrased views per target sample. Every experiment is repeated $3$ times with different seeds, and we report the mean $\pm$ std.
\paragraph{Batch Sizes.}
Dataset-specific batch sizes for \emph{full-model} LoRA fine-tuning:
\begin{table}[H]
\centering
\small
\setlength{\tabcolsep}{4pt}
\renewcommand{\arraystretch}{1.0}
\begin{tabular}{l||cccc}
\Xhline{1.2pt}
\rowcolor{tablehead!20}
\textbf{Dataset} & \textbf{RWKU} & \textbf{WMDP} & \textbf{MUSE} & \textbf{WaterDrum} \\
\hline\hline
\rowcolor{gray!10}\textbf{Batch Size} & $8$ & $6$ & $4$ & $8$ \\
\Xhline{1.2pt}
\end{tabular}
\vspace{-0.5em}
\caption{LoRA fine-tuning batch size (target 7B model).}
\label{tab:lora_batch_size_app}
\end{table}

Batch sizes for \emph{decoder} training on the proxy model:
\begin{table}[H]
\centering
\small
\setlength{\tabcolsep}{4pt}
\renewcommand{\arraystretch}{1.0}
\begin{tabular}{l||cccc}
\Xhline{1.2pt}
\rowcolor{tablehead!20}
\textbf{Dataset} & \textbf{RWKU} & \textbf{WMDP} & \textbf{MUSE} & \textbf{WaterDrum} \\
\hline\hline
\rowcolor{gray!10}\textbf{Batch Size} & $32$ & $24$ & $16$ & $32$ \\
\Xhline{1.2pt}
\end{tabular}
\vspace{-0.5em}
\caption{Decoder training batch size (proxy 3B model).}
\label{tab:decoder_batch_size_app}
\end{table}

We apply gradient accumulation during LoRA training to simulate an effective global batch size of 32 for all baselines. Decoder training uses the 1k held-out set and converges in a few hundred steps due to the lightweight architecture.

\section{Additional Experiments}\label{app:proxy-transfer}
\subsection{Transferability of proxy and target models}
In our main experiments, R2F adopts Mistral-7B as the target model and Mistral-3B as the proxy model. To further analyze the impact of proxy-target alignment, we conduct an additional experiment by varying both the target model (used for unlearning) and the proxy model (used for gradient decoding). Specifically, we evaluate R2F across three target models: Mistral-7B, GPT2-XL~\footnote{\url{https://huggingface.co/openai-community/gpt2-xl}}, and LLaMA 3.1 8B~\footnote{\url{https://huggingface.co/meta-llama/Llama-3.1-8B}}, and pair each with three proxy models of similar architectures but different scales: Mistral-3B, GPT2-Medium~\footnote{\url{https://huggingface.co/openai-community/gpt2-medium}}, and LLaMA 3.2 3B~\footnote{\url{https://huggingface.co/meta-llama/Llama-3.2-3B}}, respectively.  As shown in the tables below, R2F achieves the best performance when the proxy and target models are architecturally aligned, confirming that proxy-target similarity plays a crucial role in gradient recovery effectiveness.

\begin{table*}[h]
\centering
\small
\setlength{\tabcolsep}{6pt}
\renewcommand{\arraystretch}{1.0}
\begin{tabular}{l||ccc}
\Xhline{1.2pt}
\rowcolor{tablehead!20}
\textbf{Target $\backslash$ Proxy} & \textbf{LLaMA 3.2 3B} & \textbf{Mistral 3B} & \textbf{GPT2-Medium} \\
\hline\hline
\rowcolor{gray!10}LLaMA 3.1 8B
& $\mathbf{86.5 \pm 0.4}$ & $84.9 \pm 0.6$ & $81.7 \pm 0.6$ \\
Mistral 7B
& $87.0 \pm 0.3$ & $\mathbf{89.3 \pm 0.3}$ (R2F) & $83.8 \pm 0.4$ \\
\rowcolor{gray!10}GPT2\text{-}XL
& $82.5 \pm 0.4$ & $83.0 \pm 0.5$ & $\mathbf{85.6 \pm 0.4}$ \\
\Xhline{1.2pt}
\end{tabular}
\caption{\textbf{USR (\%) on RWKU with different target–proxy combinations.}}
\label{tab:rwku_usr_app}
\vspace{-1em}
\end{table*}

\begin{table*}[h]
\centering
\small
\setlength{\tabcolsep}{6pt}
\renewcommand{\arraystretch}{1.0}
\begin{tabular}{l||ccc}
\Xhline{1.2pt}
\rowcolor{tablehead!20}
\textbf{Target $\backslash$ Proxy} & \textbf{LLaMA 3.2 3B} & \textbf{Mistral 3B} & \textbf{GPT2-Medium} \\
\hline\hline
\rowcolor{gray!10}LLaMA 3.1 8B
& $\mathbf{94.6 \pm 0.3}$ & $93.5 \pm 0.4$ & $91.8 \pm 0.3$ \\
Mistral 7B
& $94.9 \pm 0.3$ & $\mathbf{95.7 \pm 0.2}$ (R2F) & $93.1 \pm 0.4$ \\
\rowcolor{gray!10}GPT2\text{-}XL
& $91.0 \pm 0.3$ & $91.7 \pm 0.3$ & $\mathbf{93.6 \pm 0.3}$ \\
\Xhline{1.2pt}
\end{tabular}
\caption{\textbf{GUR (\%) on RWKU with different target–proxy combinations.}}
\label{tab:rwku_gur_app}
\end{table*}

As shown in the tables, R2F achieves the best performance when the proxy and target models share the same architecture family (i.e., diagonal entries), indicating that a higher proxy–target similarity leads to better gradient decoding and unlearning effectiveness. This supports the assumption that architectural alignment is beneficial, though not strictly necessary, for R2F to perform well.

From the results in~\Cref{tab:rwku_usr_app} and \Cref{tab:rwku_gur_app}, we observe a consistent pattern across both USR and GUR metrics: performance is highest when the proxy and target models are architecturally aligned, i.e., on the diagonal of the table. For example, the R2F performance on (Mistral-7B, Mistral-3B) yields the highest USR (89.3) and GUR (95.7), outperforming other mismatched pairs. In contrast, using a proxy model with a different architecture (e.g., pairing GPT2-Medium with Mistral-7B) leads to noticeable performance drops in both forgetting and utility retention.

These results empirically support our assumption that proxy-target similarity is critical for effective gradient reconstruction. While our method can still function with mismatched architectures, the alignment between proxy and target improves both the quality of gradient approximation and the overall unlearning effectiveness.

\end{document}